\documentclass[10pt,twocolumn,letterpaper]{article}

\usepackage{subcaption}
\usepackage{float}
\usepackage[justification=justified]{caption}	
\usepackage{lscape}                                         

\usepackage[lined,ruled,linesnumbered]{algorithm2e}

\usepackage{booktabs}                   
\usepackage{multirow}

\usepackage{paralist}

\usepackage{bm}                          
\usepackage{epsfig}                      
\usepackage{times}
\usepackage{mathptmx}
\usepackage{mathtools}
\usepackage{amssymb,amsmath}   

\usepackage{units}
\usepackage{color}
\usepackage{pifont}
\usepackage{comment}

\usepackage{url}  
\usepackage[pagebackref,breaklinks,colorlinks,allcolors=cvprblue]{hyperref}
\usepackage[capitalize]{cleveref}
\crefname{section}{Sec.}{Secs.}
\Crefname{section}{Section}{Sections}
\Crefname{table}{Table}{Tables}
\crefname{table}{Tab.}{Tabs.}

\usepackage{xspace}
\usepackage{setspace}

\def\eg{e.g.,~}               
\def\ie{i.e.,~}               
\def\vs{vs.~}                 

\newlength\paramarginsize
\newlength\figmarginsize
\newlength\tabmarginsize
\newlength\secmarginsize
\newlength\figcapmarginsize
\newlength\tabcapmarginsize

\newcommand{\paramargin}{\vspace{\paramarginsize}}
\newcommand{\figmargin}{\vspace{\figmarginsize}}
\newcommand{\tabmargin}{\vspace{\tabmarginsize}}
\newcommand{\figcapmargin}{\vspace{\figcapmarginsize}}
\newcommand{\tabcapmargin}{\vspace{\tabcapmarginsize}}

\newcommand{\mpage}[2]
{
\begin{minipage}{#1\linewidth}\centering
#2
\end{minipage}
}

\newcommand{\topic}[1]
{
\paramargin\noindent \textbf{#1}
}

\newcommand{\figcaption}[2]
{
\caption{
\textbf{#1.}  
#2            
}
}

\newcommand{\secref}[1]{Section~\ref{sec:#1}}
\newcommand{\figref}[1]{Figure~\ref{fig:#1}} 
\newcommand{\tabref}[1]{Table~\ref{tab:#1}}
\newcommand{\eqnref}[1]{\eqref{eq:#1}}

\long\def\ignorethis#1{}

\def\ours{\texttt{MASH-VLM}}

\newcommand{\X}{\mathbf{X}}
\newcommand{\Q}{\mathbf{Q}}
\newcommand{\K}{\mathbf{K}}
\newcommand{\V}{\mathbf{V}}

\def\xi{\mathbf{x}_i}

\def\x{\mathbf{x}}
\def\p{\mathbf{p}}

\def\V{\mathbf{V}}
\def\M{\mathbf{M}}
\def\X{\mathbf{X}}
\def\Y{\mathbf{Y}}
\def\R{\mathbb{R}}

\def\cls{\texttt{CLS} }

\def\vllm{Video-LLM}
\def\hrope{Harmonic-RoPE}
\def\rope{RoPE}
\def\unscene{UNSCENE}

\newcommand{\softmax}{\texttt{Softmax}}

\newcommand{\mrope}{\texttt{RoPE}}
\graphicspath{{figure}, {images}, {example}}

\usepackage[pagenumbers]{cvpr} 

\definecolor{cvprblue}{rgb}{0.21,0.49,0.74}
\usepackage[pagebackref,breaklinks,colorlinks,allcolors=cvprblue]{hyperref}

\def\paperID{14904} 
\def\confName{CVPR}
\def\confYear{2025}

\title{MASH-VLM: Mitigating Action-Scene Hallucination in Video-LLMs\\
through Disentangled Spatial-Temporal Representations}

\author{
Kyungho Bae$^{1,2}$\textsuperscript{*}, 
Jinhyung Kim$^{2}$, 
Sihaeng Lee$^{2}$, 
Soonyoung Lee$^{2}$, 
Gunhee Lee$^{2}$\textsuperscript{†}, 
Jinwoo Choi$^{1}$\textsuperscript{†} \\
$^{1}$Kyung Hee University \quad
$^{2}$LG AI Research \\
{\tt\small \{kyungho.bae, jinwoochoi\}@khu.ac.kr} \\ 
{\tt\small\{jinhyung.kim, sihaelng.lee, soonyoung.lee, gunhee.lee\}@lgresearch.ai}
}
\begin{document}

\twocolumn[{
\renewcommand\twocolumn[1][]{#1}
\maketitle
\vspace{-1cm}
    \begin{center}
        \includegraphics[width=0.9\linewidth]{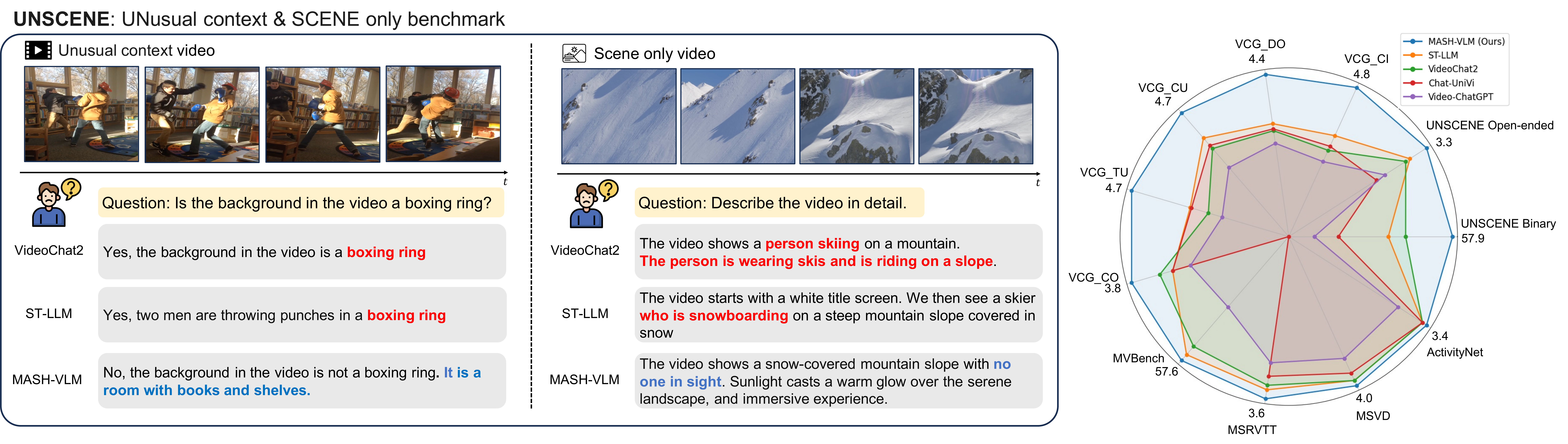}
        \vspace{-1mm}
        \mpage{0.7}{\phantom{X}\footnotesize(a)}
        \hfill
        \mpage{0.28}{{\footnotesize(b)\phantom{XXXXXXXX}}}
        
         

         \captionof{figure}{ 
         \textbf{Action-scene hallucination.}
         To evaluate action-scene hallucination in \vllm{}s, we introduce UNSCENE, an UNusual context \& SCENE-only benchmark. 
         (a) 
         When given an unusual context video of people boxing in a library, existing \vllm{}s incorrectly predict the scene as a `boxing ring’. 
         When shown a scene-only video of a snow-covered mountain with no one present, these models incorrectly identify the action as `a person skiing’ or `snowboarding on a mountain’. 
         Existing models frequently hallucinate actions based on the scene context or incorrectly predict scenes based on the observed actions. 
         (b) The proposed method, \ours{}, achieves state-of-the-art performance on the UNSCENE benchmark, as well as on existing video understanding benchmarks.
         }

        
        
        \label{fig:teaser}
    \end{center}


}]
\renewcommand{\thefootnote}{\fnsymbol{footnote}}
\footnotetext[1]{Work done during internship at LG AI Research.}
\footnotetext[2]{Corresponding authors.}

\begin{abstract}

In this work, we tackle action-scene hallucination in Video Large Language Models (Video-LLMs), where models incorrectly predict actions based on the scene context or scenes based on observed actions.
We observe that existing Video-LLMs often suffer from action-scene hallucination due to two main factors.
First, existing Video-LLMs intermingle spatial and temporal features by applying an attention operation across all tokens.
Second, they use standard Rotary Position Embedding (RoPE), which causes the text tokens to overemphasize certain types of tokens depending on their sequential orders.
To address these issues, we introduce \ours{}, Mitigating Action-Scene Hallucination in Video-LLMs through disentangled spatial-temporal representations.
Our approach includes two key innovations: 
(1) DST-attention, a novel attention mechanism that disentangles spatial and temporal tokens within the LLM by using masked attention to restrict direct interactions between spatial and temporal tokens; 
(2) Harmonic-RoPE, which extends the dimensionality of the positional IDs, allowing spatial and temporal tokens to maintain balanced positions relative to the text tokens.
To evaluate the action-scene hallucination in Video-LLMs, we introduce the UNSCENE benchmark with 1,320 videos and 4,078 QA pairs.
\ours{} achieves state-of-the-art performance on the UNSCENE benchmark, as well as on existing video understanding benchmarks.
            
\end{abstract}


%
\vspace{-7mm}
\section{Introduction}
\label{sec:intro}
\begin{figure}[t]
\centering
    
    \includegraphics[width=\linewidth]{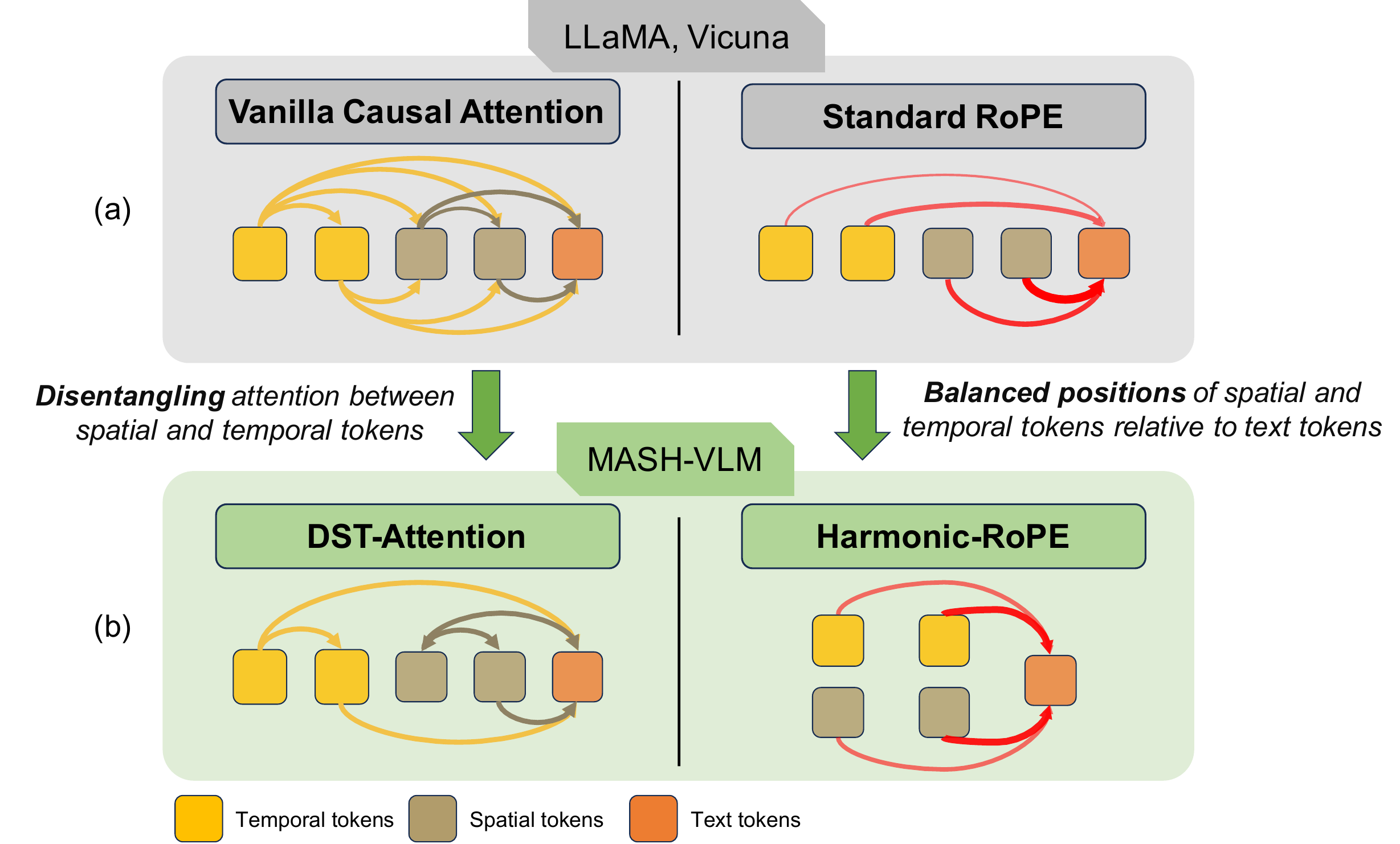}

    
    \figcaption{Comparison of attention mechanism and rotary positional embedding}
    {    
        (a) 
        In LLaMA~\cite{llama2} and Vicuna~\cite{vicuna2023}, standard causal attention among all tokens often entangles visual tokens. 
        Additionally, in standard Rotary Position Embedding (RoPE)~\cite{su2024roformer}, text tokens focus more on spatial tokens than on the temporal tokens due to the closer sequential order of spatial tokens.
        Both factors contribute to action-scene hallucinations.
        (b) 
        In contrast, our MASH-VLM employs DST-attention, where attention masking prevents direct interactions between spatial and temporal tokens, promoting feature disentanglement.
        We also introduce \hrope{}, which expands the dimensionality of standard RoPE positional IDs, allowing spatial and temporal tokens to maintain balanced positions relative to the text tokens.
        As a result, \ours{} effectively reduces action-scene hallucinations.
  }
    
    \figcapmargin
\label{fig:intro}
\end{figure}


Recent advances in natural language processing, such as in-context learning~\cite{brown2020language} and chain-of-thought reasoning~\cite{wei2022chain}, have enabled Large Language Models (LLMs) like GPT~\cite{openai2024gpt4} and LLaMA~\cite{llama2} to achieve impressive performance across various tasks.
By building on these pre-trained LLMs and applying instruction tuning~\cite{vicuna2023}, Multimodal Large Language Models (MLLMs) have become effective in integrating language with  images~\cite{liu2023llava, liu2024improvedllava1.5, li2023blip2, Zhu2024Minigpt} and videos~\cite{Maaz2023VideoChatGPT, li2023videochat,Li2024MVBench_videochat2,lin2023video-llava,zhang2023videollama, liu2025stllm}.
Despite this progress, MLLMs still face issues with hallucination, where the generated predictions are inconsistent with the visual content~\cite{biten2022let_objecthallucination,li2023pope,huang2024visual}.
We identify a specific challenge within this phenomenon: \emph{action-scene hallucination}, where models misinterpret actions based on the scene context or incorrectly infer scenes based on observed actions as shown in \figref{teaser} (a).
Given a video of a person boxing in a library, the existing Video Large Language Models (Video-LLMs) incorrectly predict the scene as a `boxing ring'.
Similarly, when shown a video of a snow-covered mountain with no one in sight, the existing Video-LLMs incorrectly predict the action as `a person is skiing’ or `snowboarding on a mountain’. 
%
%
We argue that action-scene hallucination in Video-LLMs stems from the following two limitations.

First, existing Video-LLMs often intermingle spatial and temporal features.
Without proper disentangling or debiasing techniques, this intermingling leads models to learn spurious correlations between action and scene information from videos~\cite{li2018resound,choi2019whyican,baedevias,li2023stillmix,wang2021be}.
Although some recent approaches attempt to separate spatial and temporal features~\cite{Maaz2023VideoChatGPT,xu2024slot},
they focus only on disentangling the \emph{inputs} before feeding them into an LLM.
We find that, without properly disentangling spatial and temporal features within the LLM, these features tend to re-entangle through the attention operation as shown in \figref{intro} (a), ultimately resulting in action-scene hallucination.

Second, most existing LLMs rely on the Rotary Position Embedding (RoPE)~\cite{su2024roformer} for positional encoding.
While RoPE effectively encodes positional information within a single modality, such as text, it faces limitations in multimodal contexts.
Since standard RoPE assigns 1-dimensional position IDs to input tokens based on their order, applying standard RoPE to multimodal inputs may cause a text token to attend excessively to either spatial or temporal tokens based on proximity in the sequence.
As illustrated in \figref{intro} (a), a text token might focus more on spatial tokens than on temporal tokens if spatial tokens are closer to the text token in order. 
This imbalance can lead to an over-reliance on certain token types, increasing the likelihood of action-scene hallucination~\cite{ma2024vistallama}.



To address these issues, we introduce \ours{}, a method for Mitigating Action-Scene Hallucination in Video Large Language Models through disentangled spatial-temporal representations. 
First, we propose a novel attention mechanism, DST-attention, which disentangles spatial and temporal tokens within the LLM.
As shown in \figref{intro} (b), we employ masked attention to prevent direct interactions between spatial and temporal tokens to encourage features to remain more disentangled. 
Additionally, we apply causal attention among the temporal tokens to preserve sequential dependencies and temporal order comprehension. 
Conversely, we use bi-directional attention among spatial tokens, as the spatial dimension is inherently bi-directional~\cite{dosovitskiy2020imagevit}.
Additionally, we allow the text tokens to attend to both spatial and temporal tokens to capture disentangled spatial and temporal information within the text tokens. 
By maintaining this structured attention flow, our DST-attention effectively reduces hallucinations and significantly enhances video understanding.


Second, we propose \hrope{} to overcome the limitation of standard RoPE~\cite{su2024roformer}. 
In its original form, standard RoPE does not assign equal positional IDs to spatial and temporal tokens.
To address this, we expand the dimensionality of the positional IDs, enabling spatial and temporal tokens to additionally receive balanced positional IDs relative to the text tokens, as illustrated in \figref{intro} (b).
By employing both distinct and balanced positional IDs, \hrope{} promotes a more nuanced understanding of the visual tokens, helping the model to harmoniously capture diverse spatial and temporal information.
Our empirical results show that \hrope{} is effective in mitigating action-scene hallucinations.



To evaluate action-scene hallucination in Video-LLMs, we introduce UNusual context \& SCENE-only (UNSCENE) benchmark.
The UNSCENE benchmark comprises 1,320 videos and 4,078 QA pairs, organized into two main categories: videos with unusual action-scene combinations and scene-only videos lacking any human actions.
Experimental results demonstrate that \ours{} mitigates action-scene hallucination and achieves state-of-the-art performance on the UNSCENE benchmark as well as on existing video understanding benchmarks as shown in \figref{teaser}.
We summarize our major contributions as follows:
\begin{itemize}
    \item We propose \ours{}, a method designed to mitigate action-scene hallucination in Video-LLMs with DST-attention for disentangling spatial and temporal tokens within an LLM and \hrope{} to assign balanced relative positional IDs for spatial and temporal tokens.
    \item  We introduce UNSCENE, a benchmark specifically curated to evaluate action-scene hallucination in Video-LLMs, comprising 1,320 videos and 4,078 QA pairs.
    \item \ours{} effectively alleviates action-scene hallucination, achieving state-of-the-art performance on UNSCENE benchmark and existing video QA benchmarks.
\end{itemize}
\section{Related Work}
\label{sec:related}

\topic{Multimodal LLMs for Video Understanding.}
Early works on Multimodal Large Language Models (MLLMs) primarily focus on aligning the text modality with the \emph{image} modality in the feature space~\cite{li2023blip2,liu2023llava,ye2023mplugowl,Zhu2024Minigpt}. 
More recently, there has been a growing interest in extending these models to handle the \emph{video} modality~\cite{zhang2023videollama,li2023videochat,Maaz2023VideoChatGPT,weng2024longvlm,vtime,Liu2024btadapter,jin2024chatunivi,ma2024vistallama,Li2024MVBench_videochat2,liu2025stllm,lin2023video-llava}.
Compared to image-based MLLMs (Image-LLMs), video-based MLLMs (\vllm{}) must learn temporal dynamics and manage a significantly larger volume of visual tokens spanning multiple frames.
Consequently, recent \vllm{}{} have focused on strategies such as reducing visual tokens~\cite{Li2024MVBench_videochat2,zhang2023videollama}, merging visual tokens~\cite{jin2024chatunivi}, or modeling temporal dynamics~\cite{Liu2024btadapter,lin2023video-llava}.
The most closely related work, Video-ChatGPT~\cite{Maaz2023VideoChatGPT}, employs a pooling strategy to separate spatial and temporal features before feeding them into the LLM. 
However, this approach often results in re-entangled features due to the attention mechanism within the LLM.
In contrast, we introduce DSP-attention and Harmonic-RoPE to disentangle spatial and temporal features within the LLM, enabling balanced and comprehensive video understanding.
As a result, \ours{} achieves outstanding performance on both the proposed UNSCENE benchmark and the existing video understanding benchmarks.

\topic{Hallucination in Multimodal LLMs.}
Despite the remarkable progress of MLLMs, hallucination remains a persistent and critical issue. 
In MLLMs, hallucination typically appears as the generation of textual content that misaligns with visual signals~\cite{biten2022let_objecthallucination,li2023pope,huang2024visual,zhou2023analyzing}, similar to the hallucinations observed in large language models~\cite{guerreiro2023hallucinations,bang2023multitask}.
A common form of hallucination in ImageLLMs is object hallucination~\cite{li2023pope,biten2022let_objecthallucination}, where a model incorrectly predicts the presence of objects that are not actually in the image.
To tackle image-object hallucinations, several benchmarks have been proposed~\cite{rohrbach2018objectchair,li2023pope,liu2023aligning,Guan2024hallusionbench} to evaluate how accurately ImageLLMs understand the visual content.

In contrast, hallucination in \vllm{} has received considerably less attention, even though addressing \vllm{}{} hallucination is crucial for reliable video understanding. 
Vista-LLaMA~\cite{ma2024vistallama} is a pioneering effort in addressing video hallucinations; however, it provides only qualitative results due to the lack of a benchmark for quantitative evaluation.
In this work, we identify action-scene hallucination, a prevalent form of hallucination in VideoLLMs, where models predict actions based on scene context or infer incorrect scenes based on observed actions.
To address action-scene hallucination, we propose the UNSCENE benchmark, specifically designed to evaluate action-scene hallucination in video understanding.
Additionally, we introduce a novel method, \ours{}, to effectively mitigate action-scene hallucination in Video-LLMs.

\section{Method}
\begin{figure*}[t]
\centering
    \includegraphics[width=0.9\linewidth]{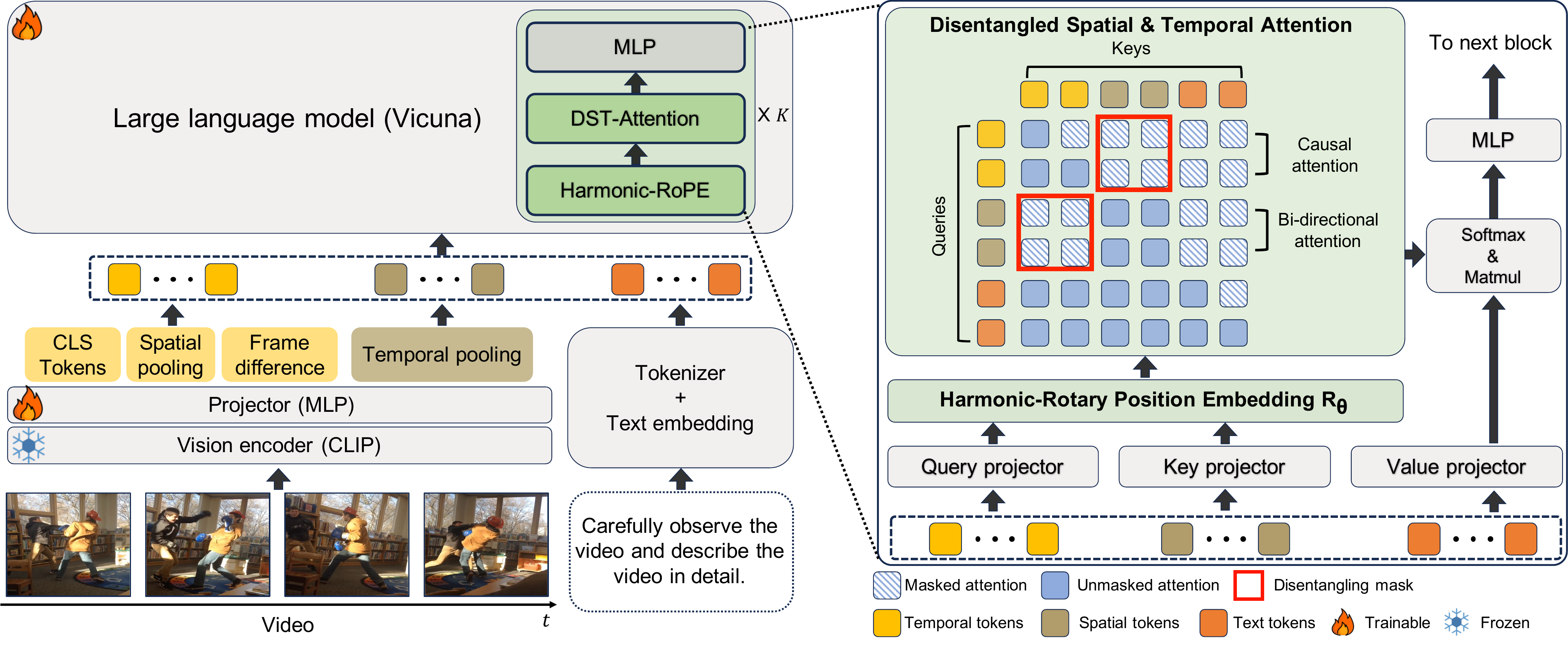}
        \figmargin
        \captionof{figure}{
        %
        \textbf{Overview of \ours{}.} 
        We propose \ours{} to mitigate action-scene hallucination. \ours{} employs Harmonic-RoPE and DST-attention within an LLM. 
        Harmonic-RoPE assigns additional \emph{balanced} positional IDs, ensuring that spatial and temporal tokens with the same ID maintain equal relative positional distances to a text token.
        DST-attention then disentangles these tokens by using masked attention, preventing direct interactions between spatial and temporal tokens.
        Together, these innovations in \ours{} effectively mitigate action-scene hallucinations and significantly enhance the model’s video understanding capabilities.
        }

\figcapmargin
\label{fig:overview}
\end{figure*}
\label{sec:method}



We introduce \ours{}, a method for Mitigating Action-Scene Hallucination in Video Large Language Models through disentangled spatial-temporal representations, as illustrated in \figref{overview}.
\ours{} comprises two key components to disentangle spatial and temporal features in an LLM: 
i) \hrope{}, which assigns balanced positional IDs to spatial and temporal tokens.
ii) DST-attention, which restricts direct attention between spatial and temporal tokens.
We begin by providing an overview of our approach in \secref{st}, followed by details on \hrope{} in \secref{harmonic} and DST-attention in \secref{dst}.





\subsection{Overview}
\label{sec:st}


Given a video $\V \in \mathbb{R}^{T \times H \times W \times C}$ with $T$ frames, spatial dimensions $H \times W$, and $C$ channels, we extract frame-level embeddings $\X \in \mathbb{R}^{T \times H/p \times W/p \times d}$ with a frame-level visual encoder, the pre-trained CLIP~\cite{clip2021Radford}, where $d$ is the embedding dimension, and $p$ is the patch size.
%
%
Next, we pass $\X$ through a learnable projection layer to obtain $\mathbf{z} \in \mathbb{R}^{T \times N \times D}$, where $N$ is the number of tokens per frame, \ie $N=H/p \times W/p$, and $D$ is the text token dimension.


\vspace{-4mm}
\paragraph{Spatial tokens.}
Given $\mathbf{z} \in \mathbb{R}^{T \times N \times D}$, we partition it along the \emph{temporal} dimension into four segments.
Next, we apply temporal pooling within each segment, followed by a $2 \times 2$ spatial pooling across the four segments to obtain the resulting \emph{spatial} tokens, $\mathbf{z}_\text{spatial} \in \mathbb{R}^{N \times D}$.


\vspace{-4mm}
\paragraph{Temporal tokens.}
Our \emph{temporal} tokens consist of i) $T$ \cls tokens of the visual encoder, ii) spatially pooled tokens, and iii) frame-difference tokens. 
Given $\mathbf{z} \in \mathbb{R}^{T \times N \times D}$, we apply \emph{spatial} average pooling to obtain the spatially pooled tokens.
To compute the frame-difference tokens, we apply spatial average pooling to the \emph{differences} between the neighboring frame tokens.
Finally, we concatenate these three types of tokens along the temporal dimension to obtain the temporal tokens $\mathbf{z}_\text{temporal} \in \mathbb{R}^{M \times D}$, where $M=3T-1$ denotes the number of temporal tokens.


Finally, we concatenate $\mathbf{z}_\text{temporal}$, $\mathbf{z}_\text{spatial}$, and text tokens $\mathbf{z}_\text{text} \in \R^{L \times D}$ along the first dimension and feed them into the LLM, where $L$ denotes the number of the text tokens.
%
%
Within the LLM, we pass the input tokens to the $K$ blocks, each consisting of \hrope{}, DST-Attention, and an MLP layer. 
We perform instruction-tuning using the prediction tokens to train \ours{}, employing standard auto-regressive training objectives~\cite{llama2,vicuna2023}.

\subsection{\hrope{}}
\label{sec:harmonic}
\begin{figure}[t]
\centering
    \includegraphics[width=0.9\linewidth]{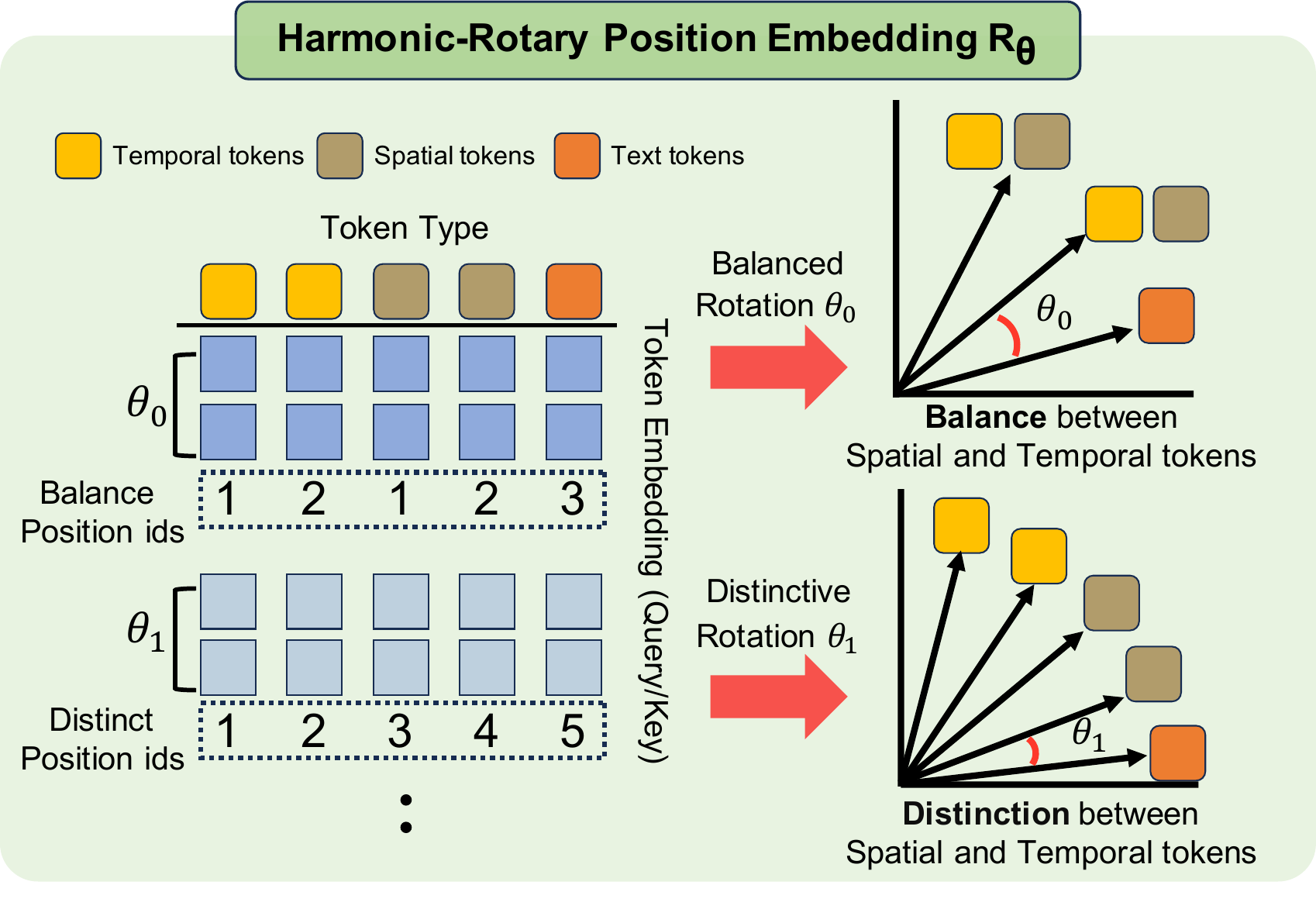}

    \figmargin
    
    \figcaption{Harmonic-RoPE}{
    %
    To overcome the limitations of standard RoPE~\cite{su2024roformer}, we propose Harmonic-RoPE. 
    In its original form, standard RoPE does not assign equal positional IDs to spatial and temporal tokens. 
    To address this, we expand the dimensionality of positional IDs, enabling spatial and temporal tokens to additionally receive balanced positional IDs relative to a text token. 
    Specifically, we assign balanced positional IDs to the even dimensions and distinct positional IDs to the odd dimensions. 
    By using Harmonic-RoPE, the model gains additional balanced positional information, leading to a more robust understanding of videos and ultimately mitigating action-scene hallucination.   
    }
    \figcapmargin
\label{fig:rope}
\end{figure}

As shown in \figref{intro} (a), applying standard RoPE~\cite{su2024roformer} to multi-modal tokens can cause the text tokens to over-focus on either the temporal or spatial tokens based on their proximity within the sequence.
Consequently, the model may become biased toward a particular token type, increasing the risk of action-scene hallucination.
To address this, we propose \hrope{}, a simple yet powerful positional embedding method that provides balanced positional information across the different token types.
By expanding the dimensionality of the positional IDs, \hrope{} enables spatial and temporal tokens to receive balanced positional IDs relative to a text token, as illustrated in \figref{rope}.

\vspace{-6mm}
\paragraph{Standard RoPE.}
Given a feature vector $\x \in \R^D$, standard RoPE rotates each 2-dimensional pair of features based on the positional ID $p$ and a predefined angle $\theta_k$ as follows:
\begin{align}
\left[\begin{array}{l}
\x^{\text{rot}}_{2k} \\
\x^{\text{rot}}_{2k+1}
\end{array}\right]
&=
\left[\begin{array}{lr}
\cos(p \theta_k) & -\sin(p \theta_k) \\
\sin(p \theta_k) & \cos(p \theta_k)
\end{array}\right]
\left[\begin{array}{l}
\x_{2k} \\
\x_{2k+1}
\end{array}\right],\\ 
\theta_k &= 10000^{-2k/D}, \quad k \in [0, D/2),
\end{align}
where $(\x_{2k},\x_{2k+1})$ denotes the $k$-th pair of features in $\x$.
By rotating each pair of features, we effectively embed relative position information within the feature representations.
$\p=[1,2,\hdots,n]$ as $\mrope(\X,\p)$ denotes the RoPE operation on the $n$ feature vectors $\X \in \R^{n \times D}$ and the position ID vector.


\vspace{-5mm}
\paragraph{Attention with \rope{}.}
To incorporate positional embedding in the attention operation, we apply \rope{} to the query and key vectors, resulting in the rotated query \(\Q^{\text{rot}} = \mrope(\Q, \p)\) and key \(\K^{\text{rot}} = \mrope(\K, \p)\).
We then use these rotated representations in the attention operation as follows:
\begin{equation}
\Y = \softmax\left(\frac{\Q^{\text{rot}} \K^{\text{rot}^T}}{\sqrt{d_k}} + \M \right) \V.
\end{equation}
This rotation directly embeds relative positional information within the attention operation, enhancing the model's ability to capture sequential dependencies.


\vspace{-4mm}
\paragraph{Distinct embeddings.}
When we have multi-modal feature vectors $\X_\text{multi} \in \R^{N+M+L}$ with $N$ spatial tokens, $M$ temporal tokens, and $L$ text tokens, we assign distinct position IDs to each token as follows:
\begin{equation}
\p_{\text{distinct}} = [1, 2, \dotsc,  N + M + L].
\label{eq:p_distinct}
\end{equation}
%
%


\vspace{-4mm}
\paragraph{Balanced embeddings.}
To address the imbalanced positional embeddings across different token types, we introduce balanced position IDs as follows:
\begin{align}
    \p_{\text{temporal}} &= [N - M + 1, \dotsc, N], \nonumber \\
    \p_{\text{spatial}} &= [1, 2, \dotsc, N], \nonumber \\
    \p_{\text{text}} &= [N + 1,N + 2, \dotsc, N + L], \nonumber \\
    \p_{\text{balanced}} &= [\p_{\text{temporal}}, \p_{\text{spatial}}, \p_{\text{text}}].
    \label{eq:p_balance}
\end{align}
Here, we assume \( N \geq  M \), as the number of spatial tokens is generally greater than the number of temporal tokens.
As illustrated in \figref{rope}, using \eqnref{p_balance} allows spatial and temporal tokens to share the same positional IDs, ensuring that both types of tokens are equi-distant from the text tokens.


\vspace{-6mm}
\paragraph{\hrope.}
In \hrope{}, we assign the balanced position IDs from \eqnref{p_balance} to the even-indexed pairs and the distinct position IDs from \eqnref{p_distinct} to the odd-indexed pairs within the feature vector $\X_\text{multi} \in \mathbb{R}^{N+M+L}$ as:
\begin{equation}
\p_\text{harmonic}(k) =
\begin{cases}
\p_{\text{balanced}}(k), & \text{if } k \text{ is even}, \\
\p_{\text{distinct}}(k), & \text{if } k \text{ is odd}.
\end{cases}
\end{equation}
Using these harmonic position IDs, we compute the rotated queries and keys as follows:
\begin{equation}
\begin{split}
\Q^{\text{rot}} = \mrope(\Q, \p_{\text{harmonic}}(k)), \\
\K^{\text{rot}} = \mrope(\K, \p_{\text{harmonic}}(k)).
\end{split}
\end{equation}
With \hrope{}, a model gains additional balanced positional information, promoting a more nuanced understanding of visual tokens and effectively capturing diverse spatial and temporal information in harmony.













\subsection{DST-Attention}
\label{sec:dst}


Existing \vllm{}s typically apply standard causal attention~\cite{li2023videochat,liu2025stllm,Li2024MVBench_videochat2,Maaz2023VideoChatGPT,weng2024longvlm,lin2023video-llava,zhang2023videollama,ma2024vistallama}, which intermingles video tokens and text tokens, often resulting in action-scene hallucinations. 
To address this, we propose the Disentangled Spatial and Temporal (DST) attention designed to prevent direct attention between spatial and temporal tokens, as illustrated in \figref{overview}. 
Specifically, we define a disentangling mask $\M^D \in \R^{(M+N) \times (M+N)}$ with \(\tau(i)\) indiciating the type of video token, either spatial or temporal token. as follows:  
\begin{equation}
    \M^{\text{D}}(i,j)= \begin{cases}
    -\infty &  \quad \text{if } \tau(i) \neq \tau(j). \\
    0 &  \quad \text{otherwise},
    \end{cases}
\end{equation}
%


To support additional spatio-temporal modeling, we introduce a structured spatio-temporal attention mask, $\M^{\text{ST}} \in \mathbb{R}^{(M+N) \times (M+N)}$.
Within the spatio-temporal attention mask, we apply causal attention among the temporal tokens to maintain sequential dependencies and temporal order comprehension.
In contrast, we employ bi-directional attention for spatial tokens, as the spatial dimension is naturally bi-directional.
We then define the overall visual attention mask as $\M^{\text{DST}} = \M^{\text{D}} + \M^{\text{ST}}.$ 
Finally, we apply causal attention to the text tokens, completing the attention mask for DST-attention.
The combination of the disentangling and structured spatio-temporal attention masks encourages features to remain more disentangled and prevents biased understanding of a video.
\begin{figure*}[t]
\centering
    \includegraphics[width=1.0\linewidth]{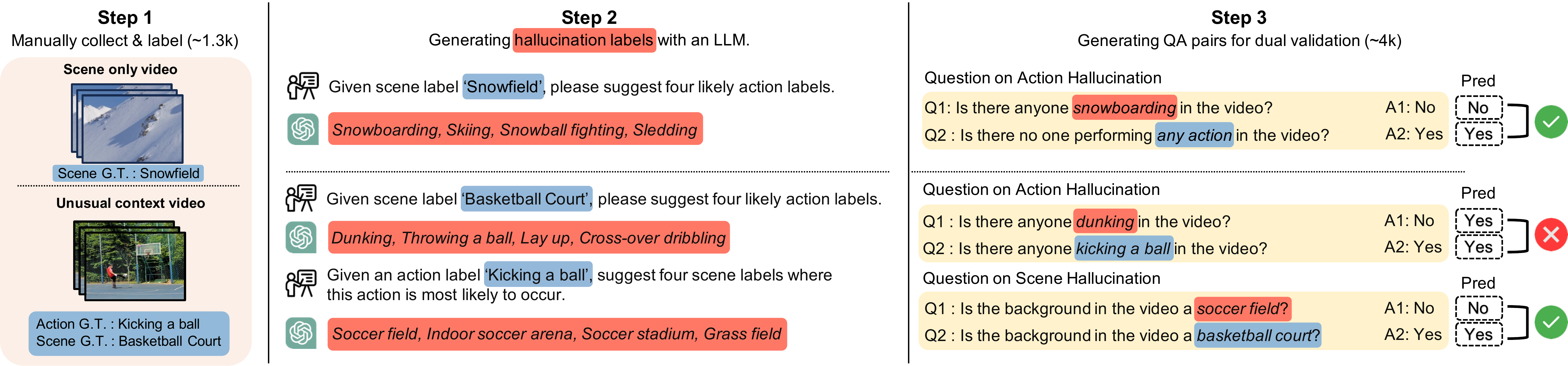}
    \vspace{-6mm}
    \captionof{figure}{
    \textbf{UNSCENE Benchmark Generation Pipeline.} 
    In step 1, we collect 1,320 videos with unusual contexts and scene-only settings from YouTube.
    In step 2, we use GPT-4~\cite{openai2024gpt4} to generate hallucination labels of which existing video-LLMs are likely to mispredict. 
    In step 3, we generate binary question-answer pairs for action and scene hallucinations using both hallucination labels and ground-truths. 
    During evaluation, a model gets a score only if it predicts both dual questions correctly.
    }
    \figcapmargin
\label{fig:pipeline}
\end{figure*}

\section{UNSCENE Benchmark}
\label{sec:overview}


To evaluate action-scene hallucination in \vllm{}s, we introduce UNSCENE, an UNusual context and SCENE-only benchmark.
We provide examples of the UNSCENE benchmark in the supplementary material.

\subsection{Benchmark}
\paragraph{Video collection.}

In \figref{pipeline}, we illustrate the process of curating the UNSCENE benchmark.
We begin by manually collecting videos from YouTube and annotating them with action and scene information.
There are two sets of videos: i) unusual context and ii) scene-only videos. 
The unusual context set consists of videos where actions are atypical for a given scene, \eg kicking a ball on a basketball court. 
The scene-only set includes videos containing only backgrounds without any people, \eg empty scenes such as a snowy mountain. 
UNSCENE comprises a total of 1,320 videos, with 719 unusual context videos and 601 scene-only videos, with an average length of 17.8 seconds. 
Notably, UNSCENE is designed solely for evaluation, not for training.

\vspace{-4mm}
\paragraph{UNSCENE binary question.}
As shown in Step 2 of \figref{pipeline}, we generate binary question-answer pairs for each video.
The QA generation process comprises two steps. 
First, we use GPT-4-turbo~\cite{openai2024gpt4} to generate plausible but \emph{incorrect} action and scene labels that are likely to cause a model to produce hallucinations.
Importantly, we do not provide GPT with any image or video input.
Instead, we prompt GPT with instructions such as, ``Given a scene label, suggest four likely action labels.'' or ``Given an action label, suggest scene labels where the action is most likely to occur.'', for generating candidates for hallucination labels.
For example, given a video of a person kicking a ball on a basketball court, the model generates hallucination labels with the action as ``Dunking, Throwing a ball, Lay up, Crossover dribbling'' and the scene as ``Soccer field, Indoor soccer arena, Soccer stadium, Grass field''.
As shown in Step 3 of \figref{pipeline}, we then curate binary QA pairs for each action and scene hallucination by randomly selecting one label from the hallucination candidates and pairing it with the ground-truth. 
For each binary QA pair, we create a question where the correct answer is ``yes'' by using the ground-truth label, and a corresponding question where the correct answer is ``no'' by using the hallucination label.
This process results in four questions per video for the unusual context set. 
Similarly, for the scene-only set, where no actions occur, we curate two action questions per video, each focused on scene details.
In total, we generate 4,078 QA pairs, comprising 2,876 pairs for the unusual context set and 1,202 pairs for the scene-only set.

\vspace{-5mm}
\paragraph{UNSCENE open-ended question.}
To evaluate action-scene hallucination in video-based conversations, we construct open-ended questions. 
We prompt a model to ``Describe the video in detail" for each of the 1,320 collected videos.
This open-ended approach enables a more comprehensive evaluation of action-scene hallucination by allowing models to generate detailed and diverse predictions.
%

\noindent
\subsection{Evaluation}
\vspace{-2mm}
\paragraph{Binary question evaluation.}

To ensure robust evaluation, we propose a \emph{dual} validation approach, requiring both questions for each video to be correctly answered for a positive score. 
For instance, a prediction is considered correct only if the model responds with ``no'' to the hallucination question and ``yes'' to the ground-truth question.

\vspace{-4mm}
\paragraph{Open-ended question evaluation.}

We employ GPT-3.5~\cite{openai_gpt3.5} to assess model responses by comparing them to the ground-truth action and scene labels. 
GPT-3.5 evaluates the model responses by scoring both action and scene aspects separately on a scale from 0 to 5, following prior work~\cite{Maaz2023VideoChatGPT}.
For scene-only videos, we assign `no action' as the ground-truth label and evaluate only the action score.








\section{Experimental Results}
\label{sec:results}




In this section, we design and conduct extensive experiments to address the following research questions:
(1) Does \ours{} more effectively mitigate \emph{action-scene hallucination} compared to state-of-the-art methods? (\secref{benchmark})
(2) How does \ours{} perform on existing video understanding benchmarks relative to current state-of-the-art approaches?  (\secref{benchmark})
(3) What is the most effective way to disentangle and balance spatial and temporal features? (\secref{ablation})
(4) Does \ours{} attend to spatial and temporal tokens as intended? (\secref{viz})
To answer these questions, we first outline the instruction tuning and benchmarks in \secref{setup}. 
For implementation details, please refer to the supplementary material.

\subsection{Experimental setup}
\label{sec:setup}
\vspace{-2mm}
\paragraph{Instruction tuning.}
Following ~\cite{Li2024MVBench_videochat2,liu2025stllm}, we utilize a diverse collection of video datasets for instruction tuning. 
Specifically, we use VideoChatGPT-100k~\cite{Maaz2023VideoChatGPT}, VideoChat-11k~\cite{li2023videochat}, Webvid~\cite{bain2021frozenwebvid}, NExT-QA~\cite{xiao2021next}, CLEVRER~\cite{yi2019clevrer}, Kinetics-710~\cite{kay2017kinetics}, and Something-Something-2~\cite{goyal2017something}, approximately 430k data points in total.
%



\begin{table*}[t]
    
\centering

\caption{\textbf{Comparison with existing methods on MVBench.}
\ours{} achieves state-of-the-art results with an average score of 57.6\%, demonstrating strong performance in general video understanding. 
The \textbf{best} and \underline{second-best} numbers are highlighted.
}

\tabcapmargin

\resizebox{\textwidth}{!}{%

\begin{tabular}{lcccccccccccccccccccccc}
\toprule
Model & LLM size & Avg & AS  & AP  & AA  & FA  & UA  & OE  & OI  & OS  & MD  & AL  & ST  & AC  & MC  & MA  & SC  & FP  & CO  & EN  & ER  & CI  \\ 
\midrule
Random & -   & 27.3 & 25.0 & 25.0 & 33.3 & 25.0 & 25.0 & 33.3 & 25.0 & 33.3 & 25.0 & 25.0 & 25.0 & 33.3 & 25.0 & 33.3 & 33.3 & 25.0 & 33.3 & 25.0 & 20.0 & 30.9 \\ 
\midrule
GPT-4V~\cite{openai2024gpt4} &Unk& 43.5 & 55.5 & \textbf{63.5} & 72.0 & 46.5 & \textbf{73.5} & 18.5 & 59.0 & 29.5 & 12.0 & \textbf{40.5} & 83.5 & \textbf{39.0} & 12.0 & 22.5 & 45.0 & \underline{47.5} & \textbf{52.0} & 31.0 & \textbf{59.0} & 11.0 \\  
Otter-V~\cite{li2023mimicotter} &7B& 26.8 & 23.0 & 23.0 & 27.5 & 27.0 & 29.5 & 53.0 & 28.0 & 33.0 & 24.5 & 23.5 & 27.5 & 26.0 & 28.5 & 18.0 & 38.5 & 22.0 & 22.0 & 23.5 & 19.0 & 19.5 \\  
mPLUG-Owl-V~\cite{ye2023mplugowl} &7B& 29.7 & 22.0 & 28.0 & 34.0 & 29.0 & 29.0 & 40.5 & 27.0 & 31.5 & 27.0 & 23.0 & 29.0 & 31.5 & 27.0 & 40.0 & 44.0 & 24.0 & 31.0 & 26.0 & 20.5 & 29.5 \\  
Video-ChatGPT~\cite{Maaz2023VideoChatGPT}  & 7B & 32.7 & 23.5 & 26.0 & 62.0 & 22.5 & 26.5 & 54.0 & 28.0 & 40.0 & 23.0 & 20.0 & 31.0 & 30.5 & 25.5 & 39.5 & \underline{48.5} & 29.0 & 33.0 & 29.5 & 26.0 & 35.5 \\  
Video-LLaMA~\cite{zhang2023videollama}  & 7B & 34.1 & 27.5 & 25.5 & 51.0 & 29.0 & 39.0 & 48.0 & 40.5 & 38.0 & 22.5 & 22.5 & 43.0 & 34.0 & 22.5 & 32.5 & 45.5 & 32.5 & 40.0 & 30.0 & 21.0 & 37.0 \\  
VideoChat~\cite{li2023videochat} & 7B & 35.5 & 33.5 & 26.5 & 56.0 & 33.5 & 40.5 & 53.0 & 40.5 & 30.0 & 25.5 & 27.0 & 48.5 & 35.0 & 20.5 & 42.5 & 46.0 & 26.5 & 41.0 & 23.5 & 23.5 & 36.0 \\  
VideoChat2~\cite{Li2024MVBench_videochat2} & 7B & 51.1 & \textbf{66.0} & 47.5 & 83.5 & \underline{49.5} & \underline{60.0} & 58.0 & \underline{71.5} & \textbf{42.5} & 23.0 & 23.0 & \textbf{88.5} & \textbf{39.0} & 42.0 & 58.5 & 44.0 & \textbf{49.0} & 36.5 & \underline{35.0} & 40.5 & \underline{65.5} \\  
ST-LLM~\cite{liu2025stllm} & 7B & \underline{54.9} &\textbf{66.0}	&53.5&	\underline{84.0}	&44.0&	58.5	&\underline{80.5}	&\textbf{73.5}	&\underline{38.5}&	\underline{42.5}	&31.0&	\underline{86.5}	&\underline{36.5}	&\underline{56.5}	&\underline{78.5}	&43.0&	44.5&	46.5	&34.5&	\underline{41.5}&	58.5 \\
\midrule
\ours{}& 7B & \textbf{57.6}&\underline{63.5}& \underline{56.0}& \textbf{89.0}& \textbf{52.0}& 56.5& \textbf{85.0}& 65.5& 37.5& \textbf{44.0}& \underline{39.0}& 83.0& 36.0& \textbf{68.5}& \textbf{89.5}& \textbf{49.5}& 45.0& \underline{47.0}& \textbf{36.5}& 39.5& \textbf{69.5}  \\
\bottomrule
\end{tabular}
}
\tabmargin
\label{tab:mvbench}
\end{table*}

\begin{table}[t]
    
\centering
\caption{\textbf{Evaluation on UNSCENE-benchmark.} 
\ours{} achieves a significant improvement over the previous state-of-the-art, with a gain of over 16\% in the binary question task and an additional 0.4-point increase in the open-ended question task out of 5 points.
The \textbf{best} and \underline{second-best} numbers are highlighted.
}
\tabcapmargin
\resizebox{1.0\linewidth}{!}{%
\begin{tabular}{lcccccccc}
\toprule
\multirow{4}{*}{Model} & \multicolumn{4}{c }{UNSCENE Binary} & \multicolumn{4}{c}{UNSCENE Open-ended} \\
\cmidrule(lr){2-5}
\cmidrule(lr){6-9}
& Scene-only & \multicolumn{2}{c }{Unusual context} & \multirow{2}{*}{Avg.} & Scene-only & \multicolumn{2}{c }{Unusual context} & \multirow{2}{*}{Avg.}  \\ 
\cmidrule(lr){2-2} 
\cmidrule(lr){3-4} 
\cmidrule(lr){6-6} 
\cmidrule(lr){7-8}
 & Action & Action & Scene  &  & Action & Action & Scene  &  \\ 
\midrule
Video-ChatGPT~\cite{Maaz2023VideoChatGPT} & 15.31 & 5.84 & 5.98 & 9.04 & 2.1 & 2.5 & 2.3 & 2.3 \\ 
VideoChat~\cite{li2023videochat} & \underline{33.61} & 22.11 & 24.20 & 26.64 & 2.7 & 2.4 & 2.7 & 2.6 \\ 
Chat-UniVi~\cite{jin2024chatunivi} & 6.66 & 22.53 & 23.37 & 17.52 & 1.7 & 2.1 & 2.5 & 2.1 \\ 
VideoChat2~\cite{Li2024MVBench_videochat2} & 30.78 & \underline{42.00} & \underline{51.04} & \underline{41.27} & 2.8 & \underline{2.7} & \underline{3.0} & 2.8 \\ 
ST-LLM~\cite{liu2025stllm} & 24.79 & \textbf{44.37} & 36.30 & 35.15 & \underline{2.9} & \textbf{2.8} & \underline{3.0} & \underline{2.9} \\ 
\midrule
\ours{} & \textbf{54.91} & 38.39 & \textbf{80.25} & \textbf{57.85} & \textbf{3.5} & \textbf{2.8} & \textbf{3.5} & \textbf{3.3} \\ 
\bottomrule
\end{tabular}
}

\tabmargin
\label{tab:halu_qa}
\end{table}

\vspace{-4mm}
\paragraph{Video understanding benchmarks.} 
We evaluate \ours{} on the \unscene{} benchmark to assess action-scene hallucination, as well as on the three public benchmarks to evaluate general video understanding capabilities.
The three public benchmarks are MVBench~\cite{Li2024MVBench_videochat2}, VCG Bench~\cite{Maaz2023VideoChatGPT}, and zero-shot video QA benchmark~\cite{Maaz2023VideoChatGPT}. 
MVBench~\cite{Li2024MVBench_videochat2} includes 20 challenging video-based tasks, each containing 200 multiple-choice samples, providing a comprehensive assessment of a model’s video comprehension abilities.
VCG-Bench~\cite{Maaz2023VideoChatGPT} comprises videos from the ActivityNet dataset~\cite{caba2015activitynet} for evaluating the quality of generated video dialogues across five distinct dimensions using GPT-3.5~\cite{openai_gpt3.5}.
The zero-shot video QA benchmark~\cite{Maaz2023VideoChatGPT} uses GPT-based evaluation on several publicly available video QA datasets, including MSRVTT-QA~\cite{xu2016msr}, MSVD-QA~\cite{xu2017msvd}, TGIF-QA~\cite{jang2019tgif}, and Activity-Net-QA~\cite{caba2015activitynet}.
Following the evaluation protocol, the GPT-3.5 model~\cite{openai_gpt3.5} scores the generated answers on a scale from 0 to 5, assessing accuracy and relevance to the ground truth.

\subsection{Comparisons with the state-of-the-art}
\label{sec:benchmark}
\paramargin
\paragraph{\unscene-Benchmark.} 
To assess the action-scene hallucination mitigation of \ours{}, we present the performance of \ours{} and existing \vllm{}s on the \unscene{} benchmark in \tabref{halu_qa}. 
In the binary QA task, \ours{} achieves state-of-the-art performance with an average score of 57.85\%, surpassing, VideoChat2, by 16.58 points. 
In the open-ended QA task, \ours{} also leads with an average score of 3.5, surpassing ST-LLM~\cite{liu2025stllm} by 0.6 points.
These results underscore the robustness of \ours{} in mitigating action-scene hallucination.


\vspace{-6mm}
\paragraph{MVBench.} 
We compare our \ours{} with previous state-of-the-art methods on MVBench. As shown in \tabref{mvbench}, our method achieves state-of-the-art results, with an average MVBench score of 57.6. \ours{} ranks best or second-best across 14 out of 20 tasks.
These results highlight that the proposed action-scene disentanglement effectively enhances the model’s ability to understand videos across various complex video tasks. 
\begin{table}[t]
\centering
\caption{
\textbf{Comparison with state-of-the-art on VCG-Bench~\cite{Maaz2023VideoChatGPT}.} 
We evaluate generated outputs across five key aspects using GPT-3.5~\cite{openai_gpt3.5}: Correctness of Information (CI), Detail Orientation (DO), Context Understanding (CU), Temporal Understanding (TU), and Consistency (CO). 
The \textbf{best} and \underline{second-best} numbers are highlighted.
}
\tabcapmargin
\resizebox{\linewidth}{!}{%

\begin{tabular}{lcccccc}
 \toprule
Method & CI   & DO   & CU   & TU   & CO   & Avg. \\  
\midrule
Video-LLaMA~\cite{zhang2023videollama}  &1.96 &2.18& 2.16& 1.82& 1.79 &1.98 \\
Video-ChatGPT~\cite{Maaz2023VideoChatGPT}  & 2.40 & 2.52 & 2.62 & 1.98 & 2.37 & 2.38 \\  
BT-Adapter~\cite{Liu2024btadapter}    & 2.68 & 2.69 & 3.27 & 2.34 & 2.46 & 2.69 \\  
VTimeLLM~\cite{vtime}    & 2.78 & \underline{3.10} & 3.40 & 2.49 & 2.47 & 2.85 \\  
Chat-UniVi~\cite{jin2024chatunivi}     & 2.89 & 2.91 & 3.46 & 2.89 & 2.81 & 2.99 \\  
LLaMA-VID~\cite{li2025llamavid}     & 2.96 & 3.00 & 3.53 & 2.46 & 2.51 & 2.89 \\  
Video-LLaVA~\cite{lin2023video-llava}    & 2.84 & 2.86 & 3.44 & 2.46 & 2.57 & 2.81 \\  
VideoChat2~\cite{Li2024MVBench_videochat2}     & 3.02 & 2.88 & 3.51 & 2.66 & 2.81 & 2.98 \\  
LongVLM~\cite{weng2024longvlm}          & 2.76	 &2.86	 &3.34 &	2.39 &	\underline{3.11} &	2.89 \\

ST-LLM~\cite{liu2025stllm}          &	\underline{3.23}	& 3.05	&\underline{3.74}	&\underline{2.93}&	2.81&	\underline{3.15} \\
\midrule
\ours{} & \textbf{4.77} & \textbf{4.38} &\textbf{4.70} & \textbf{4.67}& \textbf{3.79}& \textbf{4.46} \\

\bottomrule
\end{tabular}
}
\tabmargin
\label{tab:vcg}
\end{table}

\begin{table}[t]
\centering
\caption{\textbf{Zero-shot performance comparison with state-of-the-art on MSRVTT-QA~\cite{xu2016msr}, MSVD-QA~\cite{xu2017msvd}, TGIF-QA~\cite{jang2019tgif}, and Activity-Net-QA~\cite{caba2015activitynet}.}
The \textbf{best} and \underline{second-best} numbers are highlighted.}

\tabcapmargin

\resizebox{1.0\linewidth}{!}{%

\begin{tabular}{lcccccccc}
\toprule
\multirow{2}{*}{Method} & \multicolumn{2}{c}{MSRVTT-QA} & \multicolumn{2}{c}{MSVD-QA} & \multicolumn{2}{c}{TGIF-QA} & \multicolumn{2}{c}{ActivityNet-QA}  \\
\cmidrule(lr){2-3}
\cmidrule(lr){4-5}
\cmidrule(lr){6-7}
\cmidrule(lr){8-9}

          & Acc & Score & Acc  & Score & Acc  & Score & Acc  & Score \\  
\midrule
Video-LLaMA~\cite{zhang2023videollama}            & 29.6 & 1.8   & 51.6 & 2.5   & -    & -     & 12.4 & 1.1   \\  
VideoChat~\cite{li2023videochat}              & 45.0 & 2.5   & 56.3 & 2.8   & 34.4 & 2.3   & 26.5 & 2.2   \\  
Video-ChatGPT~\cite{Maaz2023VideoChatGPT}          & 49.3 & 2.8   & 64.9 & 3.3   & 51.4 & 3.0   & 35.2 & 2.7   \\  
Chat-UniVi~\cite{jin2024chatunivi}             & 55.0 & 3.1   & 69.3 & 3.7   & 69.0 & 3.8   & 46.1 & \underline{3.3}   \\  
LLaMA-VID~\cite{li2025llamavid} &57.7& 3.2&  69.7& 3.7& -&-& 47.4& \underline{3.3} \\
Vista-LLaMA~\cite{ma2024vistallama} & 60.5 &3.3&65.3 &3.6 &-&-&48.3& \underline{3.3} \\
Video-LLaVA~\cite{lin2023video-llava}           & 59.2 & \underline{3.5}   & 70.7 & \underline{3.9}   & \underline{70.0} & \underline{4.0}   & 45.3 & \underline{3.3}   \\  
VideoChat2~\cite{Li2024MVBench_videochat2}             & 54.1 & 3.3   & 70.0 & \underline{3.9}   & -    & -     & 49.1 & \underline{3.3}   \\  
LongVLM~\cite{weng2024longvlm}                & 59.8&	3.3	&70	&3.8	&-	&	-&47.6	&\underline{3.3} \\
ST-LLM~\cite{liu2025stllm}                 & \textbf{63.2} & 3.4   & \textbf{74.6} & \underline{3.9}   & -    & -     & \textbf{50.9} & \underline{3.3}   \\  
\midrule
\ours{}                & \underline{61.9} & \textbf{3.6}  & \underline{74.4}  & \textbf{4.0} & \textbf{72.5} & \textbf{4.1}& \underline{49.3}& \textbf{3.4} \\  
\bottomrule
\end{tabular}
}

\tabmargin
\vspace{-2mm}

\label{tab:zeroshot}
\end{table}

\begin{figure*}[t]
\centering
    \includegraphics[width=0.92\linewidth]{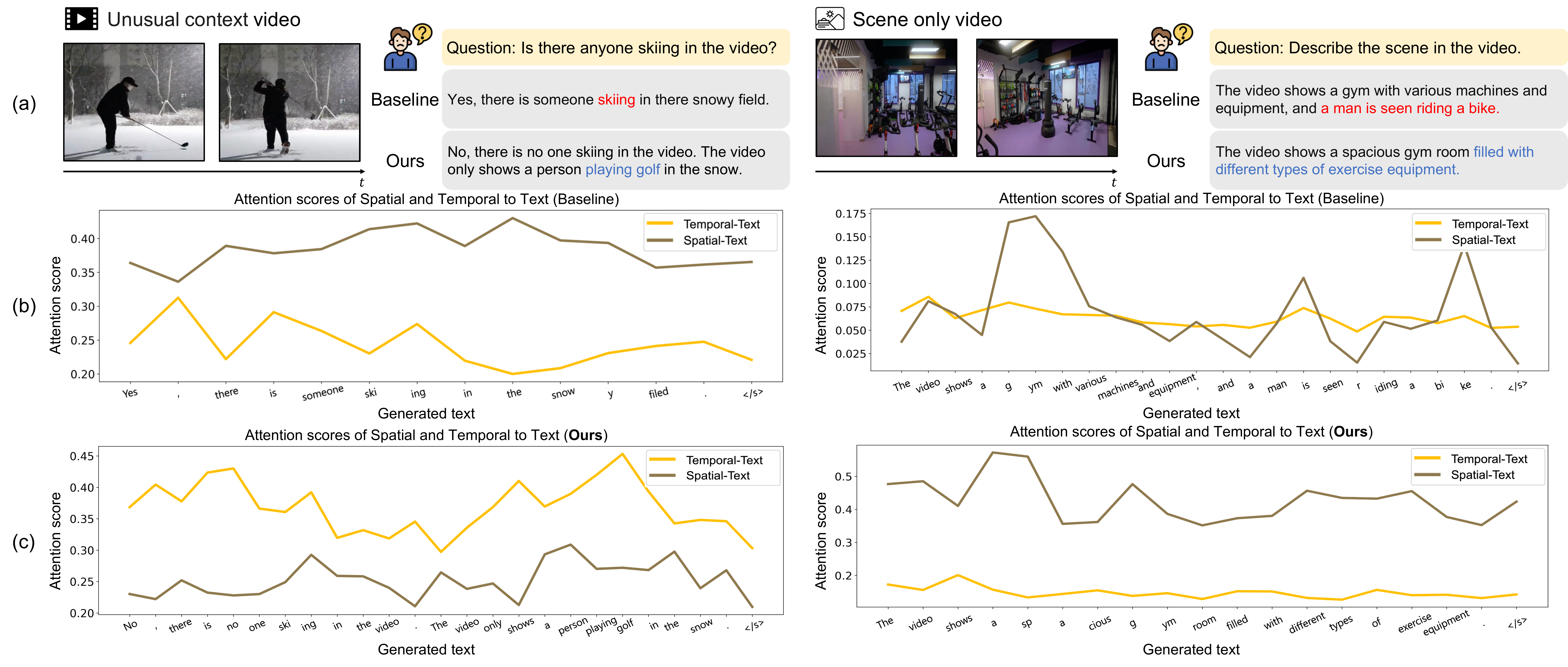}

    \vspace{-4mm}
    \figcaption{Attention scores of spatial-to-text and temporal-to-text token attention}
    {
    We compare the attention scores of the baseline model (without DST-attention and Harmonic-RoPE) and \ours{}.
    (b) The baseline model shows improper focus: when answering about actions, the model tends to focus on spatial tokens (left), and when predicting for a scene-only video, it attends to both spatial and temporal tokens (right), resulting in action-scene hallucination.
    (c) In contrast, \ours{} appropriately focuses on temporal or spatial tokens depending on the question type, effectively mitigating action-scene hallucination.
    }
    \figcapmargin
\label{fig:viz}
\end{figure*}

\vspace{-6mm}
\paragraph{VCG-Bench.} 
To assess general video understanding performance, we report the results of our \ours{} and existing methods on the VCG-Bench in \tabref{vcg}.
\ours{} significantly outperforms existing methods across all five aspects, achieving state-of-the-art performance.

\vspace{-7mm}
\paragraph{Zero-shot Video QA.} 
We evaluate zero-shot performance across several benchmark datasets in \tabref{zeroshot}. 
\ours{} achieves competitive performance across all datasets.


\noindent
\vspace{-2mm}
\subsection{Qualitative evaluation}
\vspace{-1mm}

\label{sec:viz}

To provide a deeper understanding of \ours{}, we visualize the spatial-to-text and temporal-to-text attention scores in \figref{viz}.
We show the attention scores of the baseline model (without DST-attention and \hrope{}) and \ours{}.
In \figref{viz} (b), the baseline model demonstrates improper focus: for the action QA, it tends to focus on spatial tokens (left), while for the scene QA, it focuses on both spatial and temporal tokens (right). 
The improper focus on tokens leads to action-scene hallucination as shown in \figref{viz} (a). 
In contrast, \ours{} properly focuses on temporal or spatial tokens as needed, effectively mitigating action-scene hallucination as shown in \figref{viz} (c).

\begin{table}[t]
    \centering
    \caption{\textbf{Effects of attention types for each token type.}}
    \tabcapmargin
    \resizebox{0.85\linewidth}{!}{
        \begin{tabular}{cc cc}   
        \toprule
 Temporal tokens & Spatial tokens & UNSCENE Binary & MVBench \\
 \midrule

 Causal & Causal & 41.81 & 53.41 \\
 Bi-directional & Bi-directional & 41.18 & 54.31 \\
 Bi-directional & Causal & 41.14 & 53.11 \\
 Causal & Bi-directional & \textbf{41.89} & \textbf{55.29} \\

 \bottomrule
        \end{tabular}
    }

\vspace{-2mm}
\label{tab:ablation1}
\end{table}

\begin{table}[t]
    \centering
    \caption{
    \textbf{Effect of disentangled attention between spatial and temporal tokens.}
    K$\rightarrow$Q indicates the key to query attention. 
    \checkmark represents that the corresponding attention is disabled.
    }
   \tabcapmargin

\resizebox{0.80\linewidth}{!}{
        \begin{tabular}{cccc}   
        \toprule
 Spatial $\rightarrow$ Temporal & Temporal $\rightarrow$ Spatial & UNSCENE Binary & MVBench \\
 \midrule
   \checkmark & $\times$ & 41.89 & 55.29 \\
    $\times$ & \checkmark & 42.14 & 55.09 \\
    $\times$ & $\times$ & 42.10 & 55.44 \\
   \midrule
  \checkmark & \checkmark & \textbf{45.36} & \textbf{55.51} \\

 \bottomrule
        \end{tabular}
}
\vspace{-4mm}
\label{tab:ablation2}
\end{table}



\subsection{Ablation study}
\label{sec:ablation}



\paramargin
\paragraph{Effect of the attention types.} 
We investigate the various attention mechanisms for temporal and spatial tokens in \tabref{ablation1}.
Our DST attention, which applies bi-directional attention for spatial tokens and causal attention for temporal tokens, achieves the best performance with 41.89\% on \unscene{}-Binary and 55.29\% on MVBench.
In contrast, the commonly used approach of applying causal attention to all tokens~\cite{li2023videochat,Li2024MVBench_videochat2,liu2025stllm,Maaz2023VideoChatGPT} leads to suboptimal results.


\vspace{-4mm}
\paragraph{Effect of Disentangling attention.} 
In \tabref{ablation2}, we examine the efficacy of disentangling attention in the proposed DST-attention.
Compared to the baseline without disentangling attention, using the proposed disentangling mask results in a significant performance improvement on the \unscene{}-Binary (42.10\% \vs 45.36\%).
These results indicate that restricting direct interactions between spatial and temporal tokens mitigates action-scene hallucination.


\begin{table}[t]
    \centering
    \caption{\textbf{Ablation study on positional embeddings.}}
    \tabcapmargin
    \resizebox{0.70\linewidth}{!}{
        \begin{tabular}{l cc}   
        \toprule
 Embeddings  & UNSCENE Binary & MVBench \\
 \midrule

    Distinct only & 45.36 & 55.51 \\
    Balanced only & 50.80 & 56.39 \\
    Distinct \& Balanced & \textbf{57.85} & \textbf{57.58} \\

 \bottomrule
        \end{tabular}
    }

\tabmargin
\vspace{-2mm}
\label{tab:ablation3}
\end{table}




\vspace{-4mm}
\paragraph{Effect of \hrope{}.}
In \tabref{ablation3}, we analyze the impact of \hrope{}. 
Using the balanced position embeddings alone improves performance on \unscene{}-Binary by 5.4 points compared to using distinct position embeddings \ie standard RoPE~\cite{su2024roformer}.
With both distinct and balanced position embeddings we observe a 12.5-point improvement on \unscene{}-Binary and a 2-point improvement on MVBench.
These results demonstrate that \hrope{} is effective in reducing action-scene hallucination and enhancing overall video understanding.



\vspace{-2mm}
\section{Conclusions}
\label{sec:conclusions}
\vspace{-2mm}





In this work, we introduce \ours{}, a video large language model designed to mitigate action-scene hallucination. 
\ours{} applies balanced positional embeddings through \hrope{} and disentangling spatial and temporal tokens with DST-Attention.
To evaluate the action-scene hallucination in \vllm s, we introduce the \unscene{} benchmark. 
Extensive experiments demonstrate that \ours{} effectively mitigates action-scene hallucination and achieves state-of-the-art performance on general video understanding benchmarks.

\paragraph{Acknowledgment.}
This work was supported in part by the Institute of Information \& Communications Technology Planning \& Evaluation (IITP) grant funded by the Korea Government (MSIT) under grant RS-2024-00353131, RS-2021-II212068 (Artificial Intelligence Innovation Hub), and RS-2022-00155911 (Artificial Intelligence Convergence Innovation Human Resources Development
(Kyung Hee University)).
Additionally, it was supported by the National Research Foundation of Korea (NRF) grant funded by the Korea Government (MSIT) (No. 2022R1F1A1070997) and LG AI Research.

{
    \small
    \bibliographystyle{ieeenat_fullname}
    \bibliography{main}
}

\clearpage
\section*{Supplementary Material}


    
In this supplementary material, we present implementation details, additional ablation studies, further qualitative evaluations, and examples from the UNSCENE benchmark to complement the main paper. The supplementary material is organized as follows:

\begin{enumerate}
    \setcounter{enumi}{6} 
    \item Implementation details
    \item Additional ablation studies
    \item Further qualitative evaluations
    \item Examples from the UNSCENE benchmark 
    
\end{enumerate}


\section{Implementation details}
In this study, 
we use CLIP-ViT-Large-Patch14-336~\cite{clip2021Radford} as a vision encoder and a two-layer MLP as a projector. 
The visual encoder, projector, and LLM weights are initialized with the pre-trained weights of LLaVA v1.5~\cite{liu2024improvedllava1.5}, which employs Vicuna-v1.5~\cite{vicuna2023} with 7B parameters as the language model.
During instruction tuning, we freeze the vision encoder, allowing only the projector and the LLM to be fully fine-tuned.
Following prior work~\cite{liu2024improvedllava1.5}, we set the learning rate to $2 e^{-5}$, the total batch size to $128$, and train the model for 2 epochs. 
We adopt a cosine decay learning rate schedule with a warmup ratio of $0.03$, AdamW~\cite{loshchilov2017decoupled} as the optimizer with no weight decay, and DeepSpeed Stage 3.



\section{Additional ablation study}

\paragraph{Ablation study on spatial and temporal features for video.} In \tabref{table8}, we conduct an ablation study on various video features. When using both temporal and spatial features during training, our \ours{} achieves an accuracy of 41.81\% on UNSCENE Binary and 53.41\% on MVBench, demonstrating a significant improvement over using only one of these features. Furthermore, incorporating CLS tokens and frame-difference tokens into the temporal features enhances performance, highlighting the effectiveness of our proposed feature extraction method.



\paragraph{Ablation study on LLM Tuning Scheme.} 
\tabref{table9} presents the performance results across different LLM tuning schemes. 
Full fine-tuning achieves the best performance, with accuracy of 57.85\% on UNSCENE Binary and 57.78\% on MVBench, outperforming both freezing LLM parameters and using LoRA~\cite{hu2022lora} tuning.

\paragraph{Effect of token types during inference.}
In \tabref{table10}, we investigate whether the spatial and temporal tokens of \ours{} preserve their respective information in a disentangled manner.
When only spatial tokens are used during inference, the performance on action-related QA for unusual context videos, which require temporal understanding, drops by 11.24 points. 
Conversely, using only temporal tokens during inference results in a performance decrease of 23.64 points on scene-related QA for scene-only videos and 18.1 points for unusual context videos, both of which require spatial understanding.
These results demonstrate that the disentangled tokens effectively preserve their respective information: spatial tokens retain spatial details, while temporal tokens capture temporal dynamics. 
Furthermore, by leveraging both disentangled tokens, \ours{} achieves the highest performance.

\begin{table}[t]
    \centering

    \caption{\textbf{Ablation study on spatial and temporal features for video.} 
    SP, TP, and F-Diff refer to spatial pooling, temporal pooling, and frame difference, respectively.
    }\hfill
    
\mpage{1.0}{
    \centering
    \resizebox{\linewidth}{!}{
    \begin{tabular}{cccc}   
    \toprule
 Temporal token & Spatial tokens &  UNSCENE Binary & MVBench \\
 \midrule 

  \ding{55} & TP  & 9.94 & 37.15 \\
 SP & \ding{55}  & 28.19 & 45.98 \\
 SP $+$ CLS & \ding{55}  &  30.30 & 48.02 \\
 SP $+$ CLS  $+$ F-Diff & \ding{55}  & 31.29 & 48.77 \\
 SP $+$ CLS  $+$ F-Diff &  TP  & \textbf{41.81} & \textbf{53.41} \\
 \bottomrule
    \end{tabular}
}
}

    \label{tab:table8}
\end{table}

\begin{table}[t]
    \centering
    \caption{\textbf{Ablation study on LLM Tuning Scheme.}}

\mpage{1.0}{
    \centering
    \resizebox{0.8\linewidth}{!}{
    \begin{tabular}{ c cc}   
    \toprule
 LLM Tuning&  UNSCENE Binary & MVBench \\
 \midrule
   Frozen& 37.49 & 48.41  \\
LoRA& 44.31 & 51.10 \\
  Full F.T.& \textbf{57.85} & \textbf{57.78}  \\

 \bottomrule

    \end{tabular}
}
}

    \label{tab:table9}
\end{table}

\begin{table}[t]
    \centering
    \caption{\textbf{Effect of token types during inference.}}

\mpage{1.0}{
    \centering
    \resizebox{0.8\linewidth}{!}{
    \begin{tabular}{ c ccc}   
    \toprule
 \multirow{2}{*}{\raisebox{-2mm}{\shortstack{Video token types}}} & Scene-only & \multicolumn{2}{c }{Unusual context}\\
 
 \cmidrule(lr){2-2}
 \cmidrule(lr){3-4}
 & Scene & Action & Scene \\

 \midrule
   Spatial& 49.75 & 27.15 & 74.85 \\
Temporal& 31.27 & 33.13 & 62.15  \\
  Spatial \& Temporal & \textbf{54.91} & \textbf{38.39} & \textbf{80.25}   \\

 \bottomrule

    \end{tabular}
}
}

\vspace{-4mm}
    \label{tab:table10}
\end{table}

\begin{figure}[h]
\centering
    \includegraphics[width=0.96\linewidth]{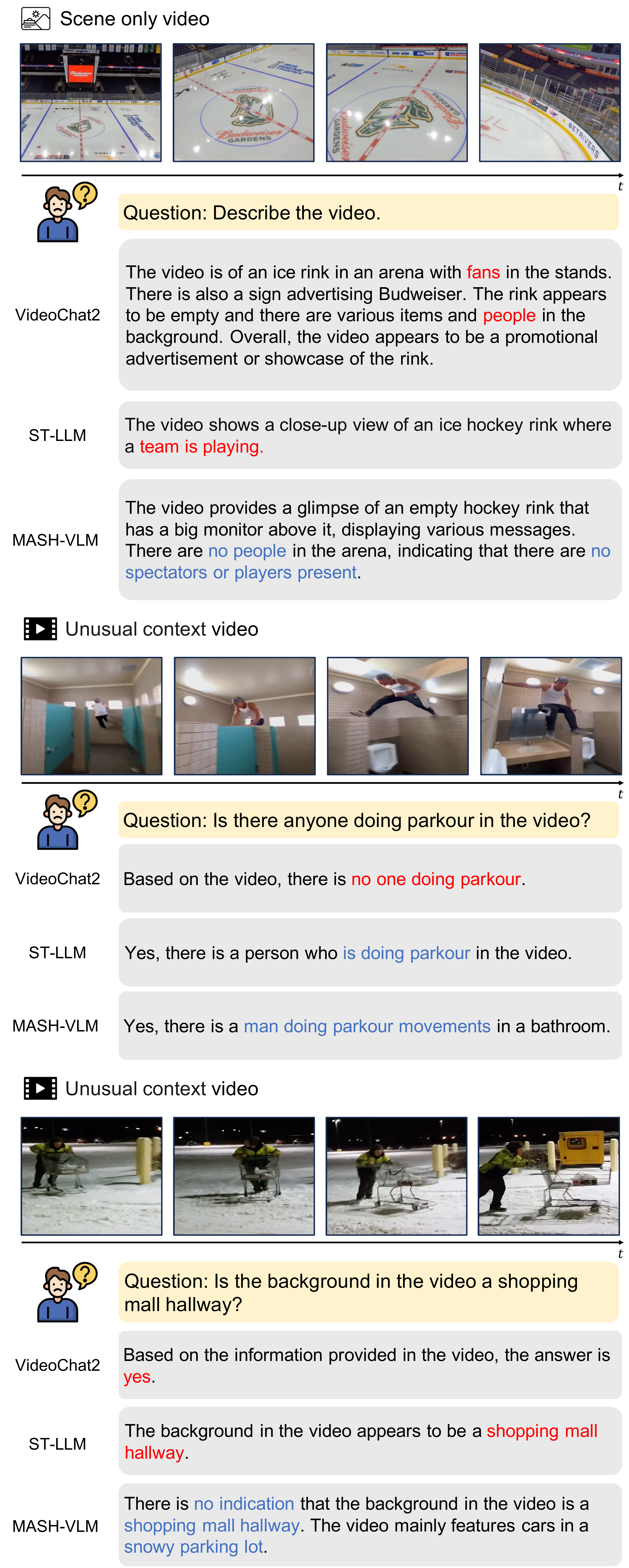}
    
    \figcaption{Qualitative results on the UNSCENE benchmark}
    {
    }

    \vspace{-2mm}
\label{fig:viz_qual}
\end{figure}

\section{Further qualitative evaluations}

\paragraph{Qualitative results.}
In \figref{viz_qual}, we present qualitative comparisons with other methods.
In the top example, a scene-only video depicting an ice hockey rink without any people is shown. 
Previous video-LLMs incorrectly respond that people are present in the background or that a team is playing a game. 
In contrast, \ours{} not only accurately predicts the absence of people but also provides a detailed description of the background.
In the middle example, an unusual context video shows a person performing parkour in a restroom. 
While VideoChat2~\cite{Li2024MVBench_videochat2} incorrectly predicts that no one is performing parkour, both ST-LLM~\cite{liu2025stllm} and \ours{} correctly identify the presence of a person engaged in parkour.
Similarly, as shown in the bottom example, \ours{} accurately identifies the background.
These qualitative results demonstrate that \ours{} effectively mitigates action-scene hallucinations.
%

\begin{figure*}[t]
\centering
    \includegraphics[width=1.0\linewidth]{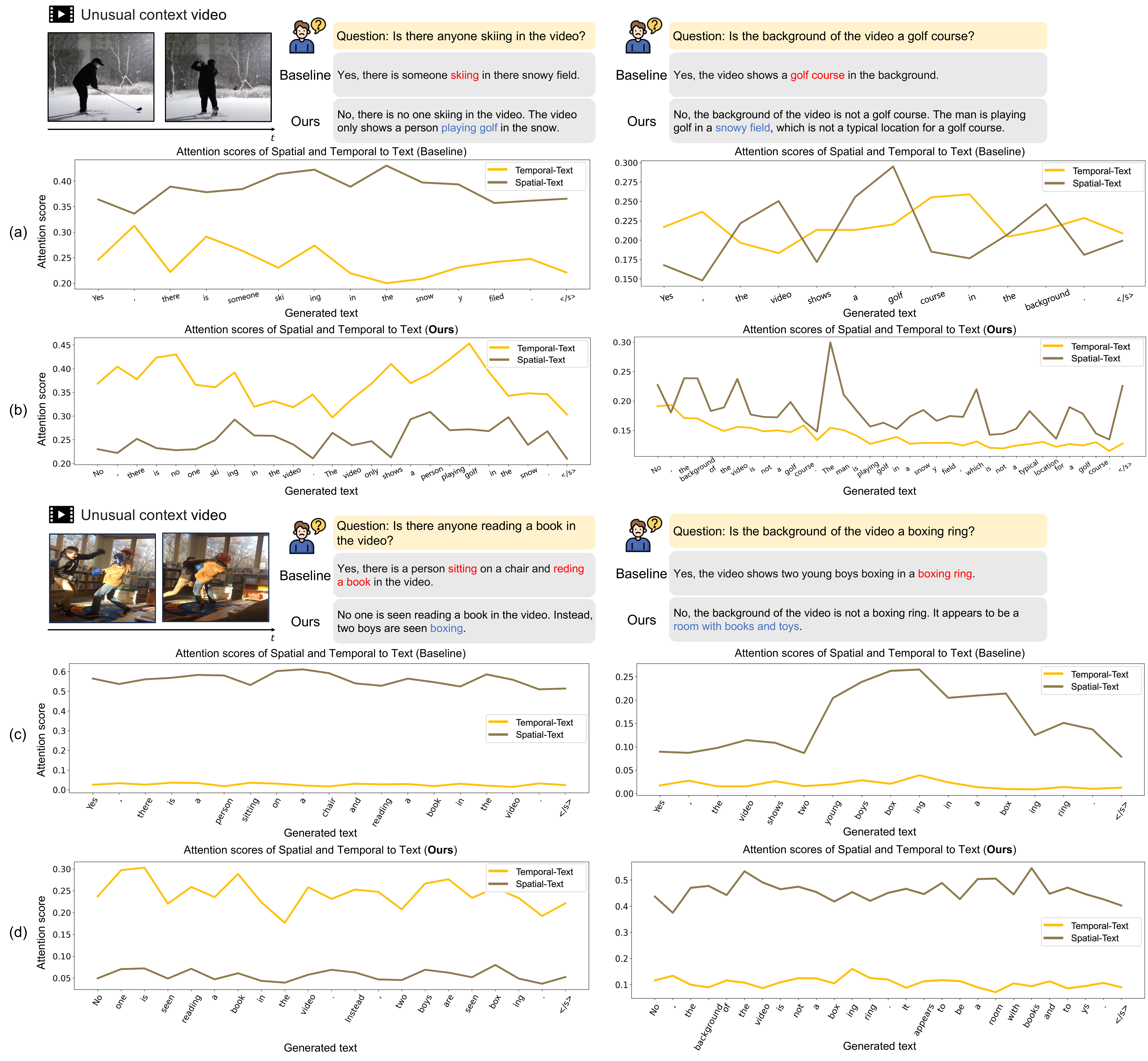}
    
    \figcaption{Attention scores of spatial-to-text and temporal-to-text token attention}
    {
We compare the attention scores of the baseline model (without DST-attention and Harmonic-RoPE) and \ours{}.
(a) The baseline model shows improper focus: when answering about actions, the model tends to focus on spatial tokens (left), and when answering about scenes, it attends to both spatial and temporal tokens (right), resulting in action-scene hallucination.
(b) \ours{} appropriately focuses on temporal or spatial tokens depending on the question type.
(c) When answering about actions, the baseline model tends to focus on spatial tokens (left), and when answering about scenes, it attends to spatial tokens but generates a hallucinated response (right).
This hallucination arises from the baseline model’s failure to disentangle spatial and temporal tokens, leading to entanglement between spatial and temporal tokens.
(d) In contrast, \ours{} not only focuses on temporal or spatial tokens depending on the question type but also learns disentangled spatial and temporal representation, effectively mitigating action-scene hallucination.
    }
\label{fig:viz_sup}
\end{figure*}

\vspace{-4mm}
\paragraph{Attention scores.}
We compare the attention scores of the baseline model (without DST-attention and Harmonic-RoPE) and \ours{} as shown in \figref{viz_sup}.
(a) The baseline model shows improper focus: when answering about actions, the model tends to focus on spatial tokens (left), and when answering about scenes, it attends to both spatial and temporal tokens (right), resulting in action-scene hallucination.
(c) When answering about actions, the baseline model tends to focus on spatial tokens (left), and when answering about scenes, it attends to spatial tokens but generates a hallucinated response (right).
This hallucination arises from the baseline model’s failure to disentangle spatial and temporal tokens, leading to entanglement between spatial and temporal tokens.
(b,d) In contrast, \ours{} appropriately focuses on temporal or spatial tokens depending on the question type and learns disentangled spatial and temporal representation, effectively mitigating action-scene hallucination.

\section{Examples of UNSCENE benchmark}
\begin{figure*}[t]
\centering
    \includegraphics[width=0.95\linewidth]{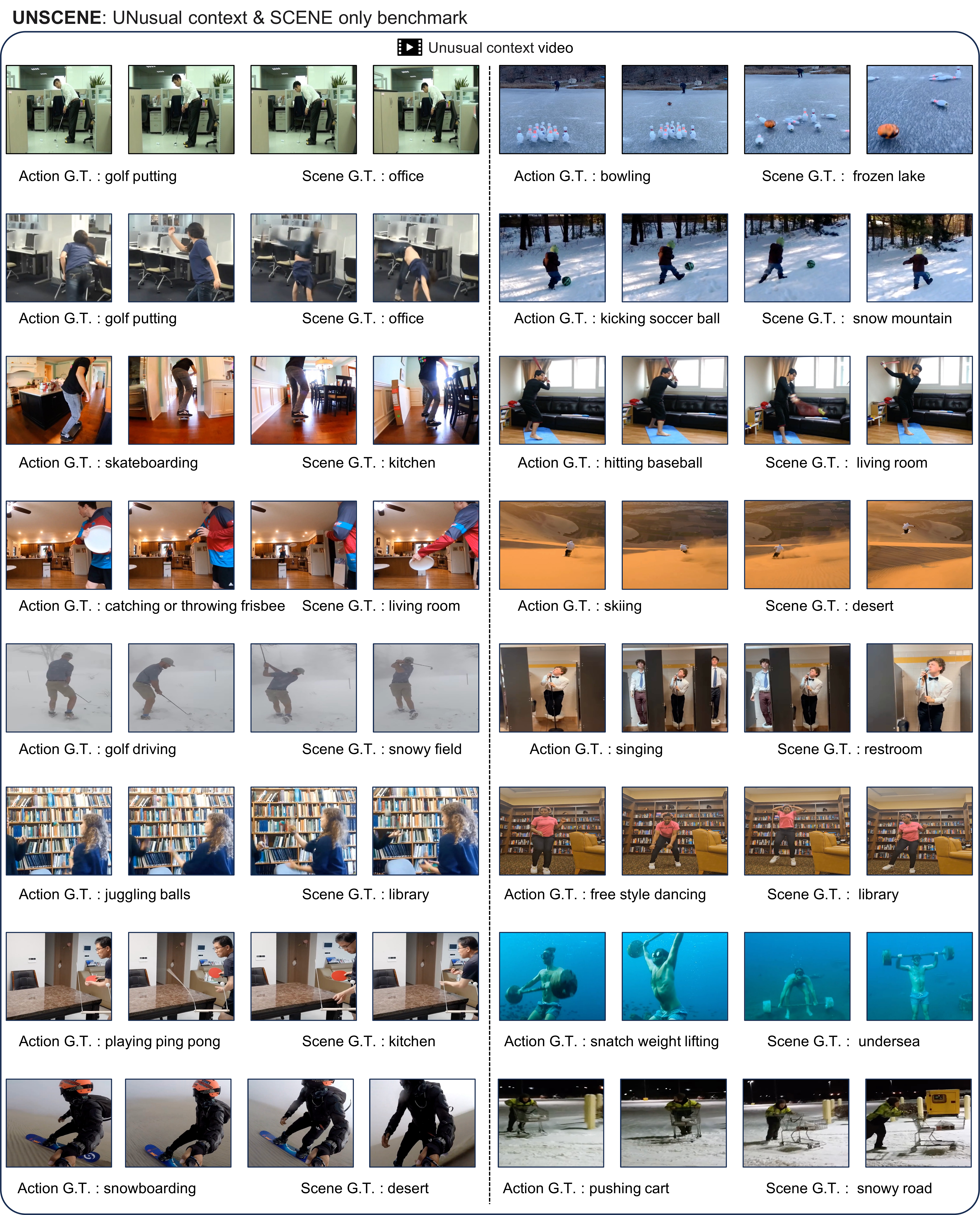}

    \caption{\textbf{Examples of unusual context videos in the UNSCENE benchmark.}}
    \label{fig:mis1}
\end{figure*}

\begin{figure*}[t]
\centering
    \includegraphics[width=0.95\linewidth]{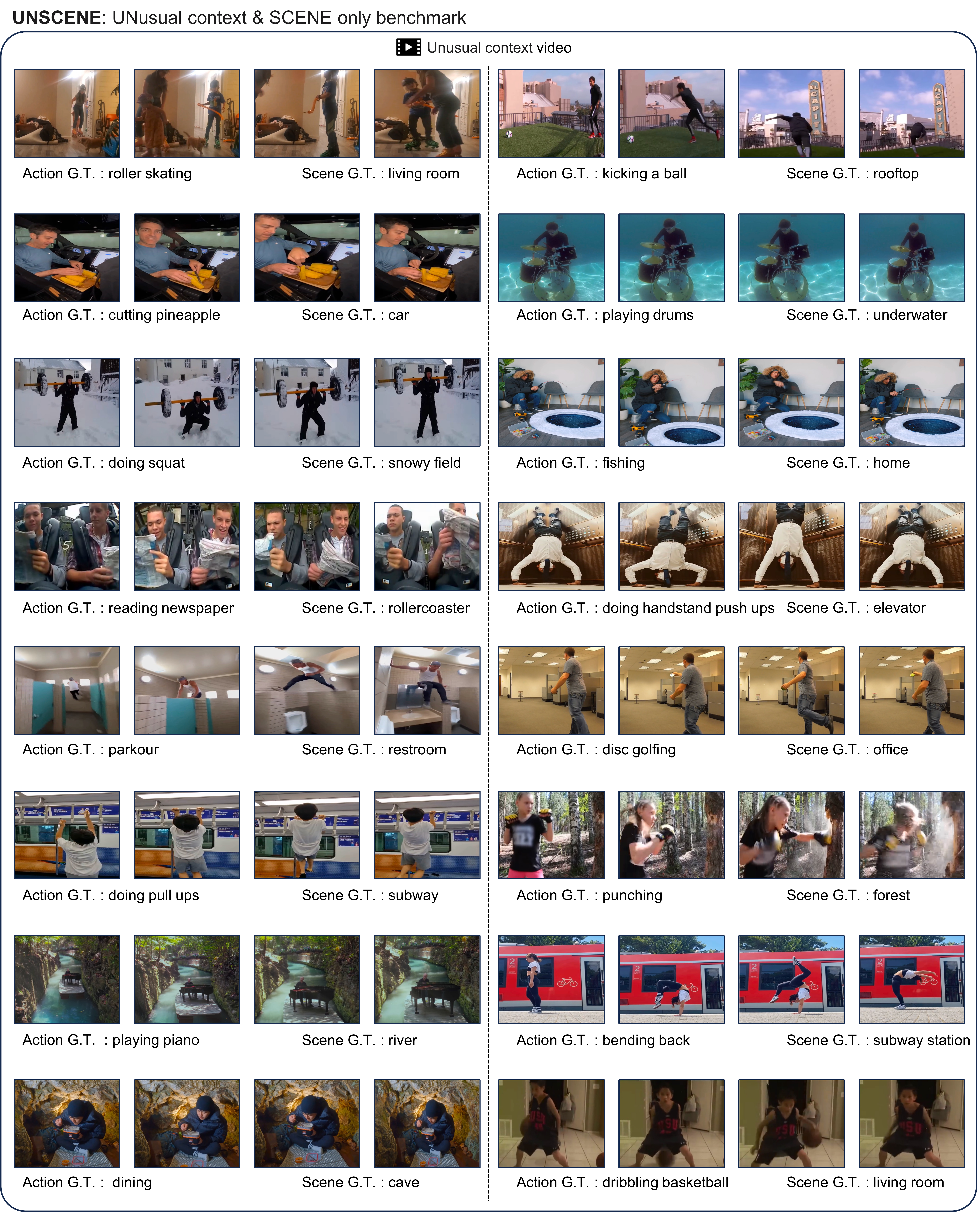}

    \caption{\textbf{Examples of unusual context videos in the UNSCENE benchmark.}}
    \label{fig:mis2}
\end{figure*}

In \figref{mis1} and \figref{mis2}, we showcase examples of unusual context videos in the UNSCENE benchmark. We also present examples of scene-only videos as shown in \figref{so}. We also provide example videos of UNSCENE benchmark in the supplementary material.

\begin{figure*}[t]
\centering
    \includegraphics[width=0.95\linewidth]{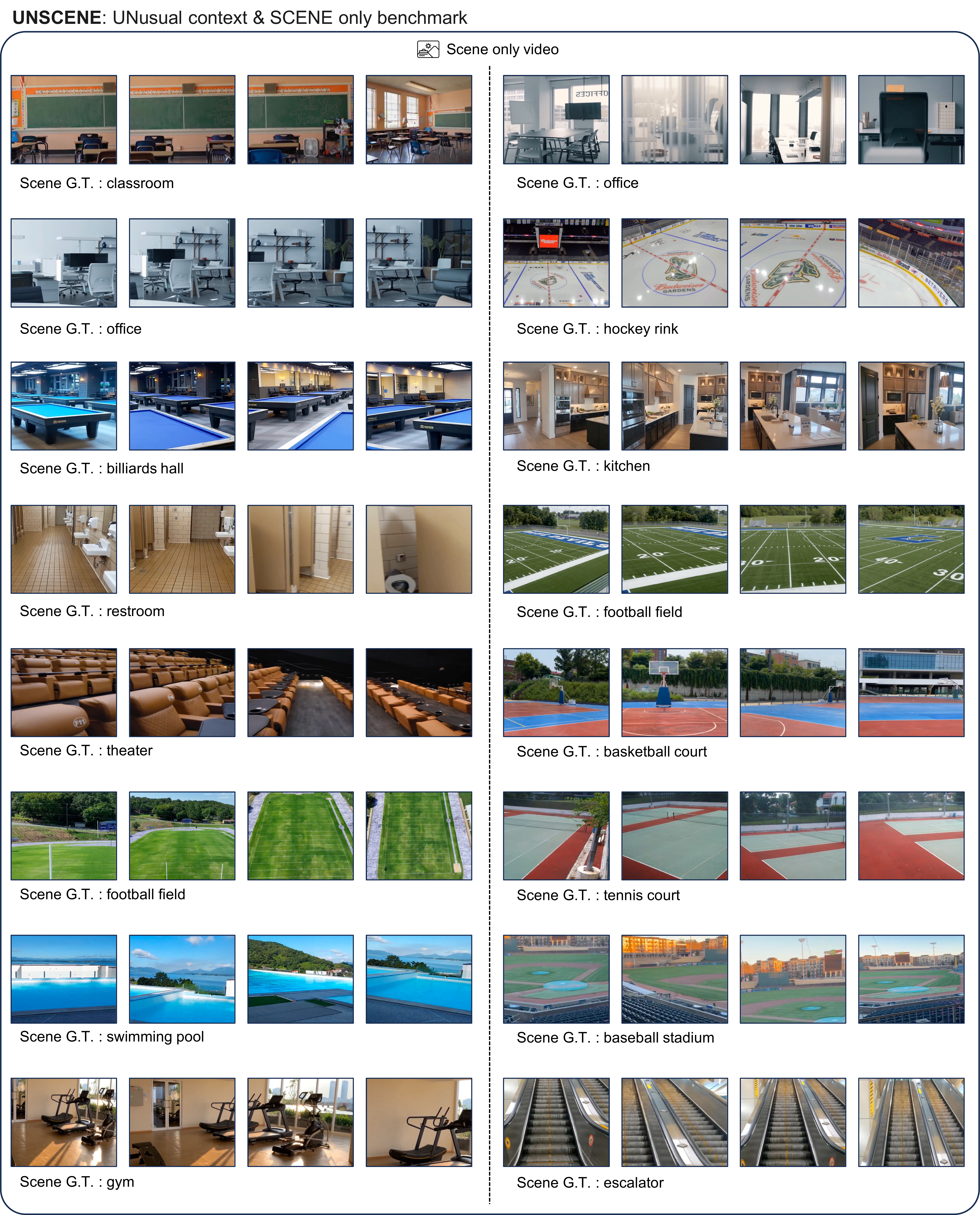}

    \caption{\textbf{Examples of scene only videos in the UNSCENE benchmark.}}
\label{fig:so}
\end{figure*}


\end{document}